\documentclass[10pt,twocolumn,letterpaper]{article}

\usepackage[pagenumbers]{cvpr} 

\usepackage{graphicx}
\usepackage{amsmath}
\usepackage{amssymb}
\usepackage{booktabs}

\usepackage[pagebackref,breaklinks,colorlinks]{hyperref}

\usepackage[capitalize]{cleveref}
\crefname{section}{Sec.}{Secs.}
\Crefname{section}{Section}{Sections}
\Crefname{table}{Table}{Tables}
\crefname{table}{Tab.}{Tabs.}

\usepackage{color}
\usepackage{kotex}
\usepackage{multirow}
\usepackage{booktabs}
\usepackage{float}
\usepackage{adjustbox}
\usepackage{xurl}

\definecolor{darkergreen}{RGB}{21, 152, 56}
\definecolor{red2}{RGB}{252, 54, 65}
\newcommand\redp[1]{\textcolor{red2}{(#1)}}
\definecolor{iblue}{rgb}{0.0,0.44,0.75}
\newcommand\greenp[1]{\textcolor{darkergreen}{(#1)}}
\DeclareMathOperator*{\argmin}{argmin}
\DeclareMathOperator*{\argmax}{argmax}
\def\cvprPaperID{9} 
\def\confName{CVPR}
\def\confYear{2022}

\newcommand\blfootnote[1]{%
  \begingroup
  \renewcommand\thefootnote{}\footnote{#1}%
  \addtocounter{footnote}{-1}%
  \endgroup
}

\begin{document}

\title{RewriteNet: Reliable Scene Text Editing with \\ Implicit Decomposition of Text Contents and Styles}

\author{Junyeop Lee$^{1*}$, Yoonsik Kim$^{2*}$ \and Seonghyeon Kim$^2$, Moonbin Yim$^2$, Seung Shin$^2$, Gayoung Lee$^2$, Sungrae Park$^{1\dagger}$\vspace{1.0em} \\ 
$^1$Upstage AI Research, Upstage \\
$^2$Clova AI Research, NAVER Corp.\\
\tt\small\{junyeop.lee, sungrae.park\}@upstage.ai \\
\tt\small\{yoonsik.kim90, kim.seonghyeon, moonbin.yim, seung.shin, gayoung.lee\}@navercorp.com
}


\maketitle

\begin{abstract}
Scene text editing (STE), which converts a text in a scene image into the desired text while preserving an original style, is a challenging task due to a complex intervention between text and style. 
In this paper, we propose a novel STE model, referred to as RewriteNet, that decomposes text images into content and style features and re-writes a text in the original image. Specifically, RewriteNet implicitly distinguishes the content from the style by introducing scene text recognition. Additionally, independent of the exact supervisions with synthetic examples, we propose a self-supervised training scheme for unlabeled real-world images, which bridges the domain gap between synthetic and real data.
Our experiments present that RewriteNet achieves better generation performances than other comparisons. 
Further analysis proves the feature decomposition of RewriteNet and demonstrates the reliability and robustness through diverse experiments. 
Our implementation is publicly available at \url{https://github.com/clovaai/rewritenet}

\end{abstract}
\blfootnote{* indicates equal contribution.}
\blfootnote{$\dagger$ indicates corresponding author.}
\section{Introduction}
\begin{figure}
  \centering
  \begin{subfigure}{0.98\linewidth}
    \includegraphics[width=1.0\linewidth]{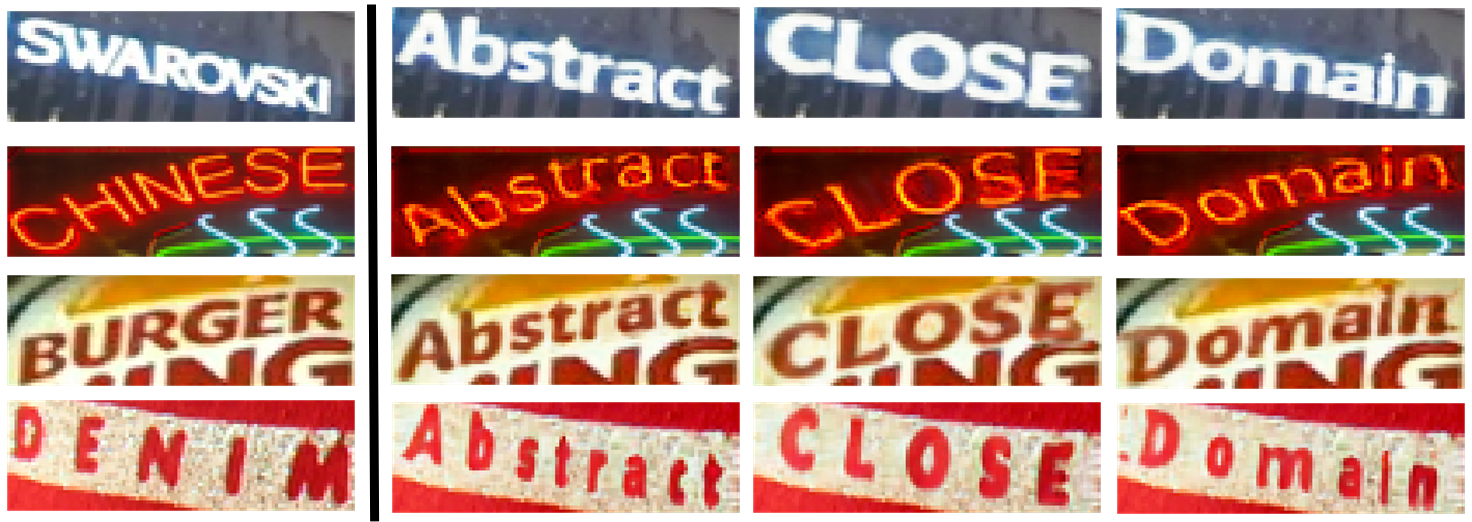}
    \caption{Original and edited text images.}
    \label{fig:ex_rn}
  \end{subfigure}
  \hfill
   \begin{subfigure}{0.98\linewidth}
    \includegraphics[width=1.0\linewidth]{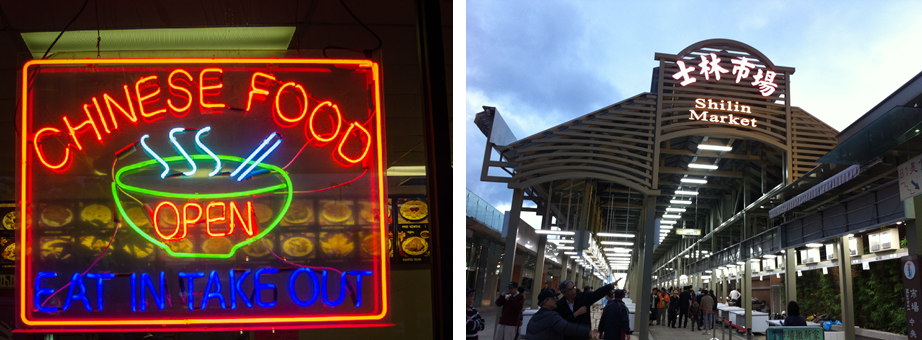}
    \caption{Original scene images.}
    \label{fig:ex_original}
  \end{subfigure}
  \hfill
   \begin{subfigure}{0.98\linewidth}
    \includegraphics[width=1.0\linewidth]{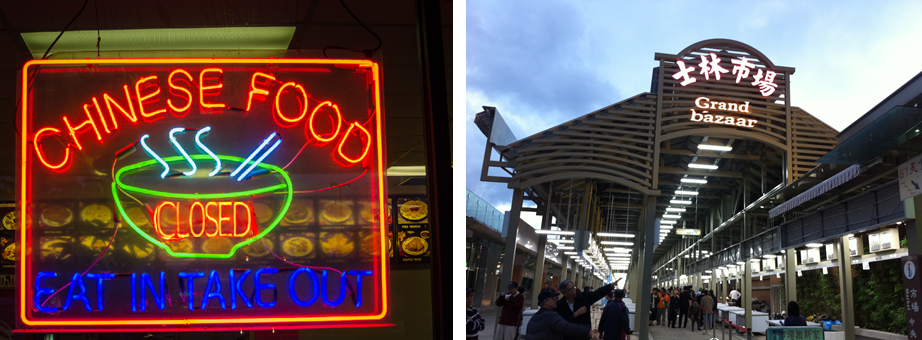}
    \caption{Edited scene images.}
    \label{fig:ex_edited}
  \end{subfigure}
   \caption{Examples of STE results. (a) is the original text images (leftmost) and the text edited images where target texts are ``Abstract'', ``CLOSE'', and ``Domain''. (b) and (c) show real-world applications with original and edited scene images. All results are generated by RewriteNet.}
    \label{fig:teaser}
\end{figure}


Scene text editing (STE) is a task of image synthesis that replaces the text in a scene image to the desired text while preserving a style such as a font type, font size, text alignment, and background. As shown in Figure ~\ref{fig:teaser} (a), the texts in the image patches are converted while keeping the original styles. 
As a core technology for virtual reality, STE can be employed for scene text images to replace the text contents (e.g. Figure~\ref{fig:teaser} (b) and (c)).
As can be seen, since real-world text images have complex backgrounds and text styles, STE methods should address intricately intertwined tasks including image in-painting, style extraction, character rendering, and text localization. 


Previous STE methods~\cite{stefann2020,wu2019etiw,swaptext2020} follow 
a framework with two stages: text deletion and text conversion.
Text deletion module generates text erased background, which can be thought of as an image in-painting task specialized for scene text images~\cite{nakamura2017scene,yu2018generative,xiong2019foreground,tursun2019mtrnet}.
Text conversion module renders the desired text where the text-related styles in the original image are transferred, and then, two outputs generated from text deletion and conversion module are harmonically fused.
By incorporating the text deletion, previous methods show the feasibility of STE. However, since the text deletion heavily depends on visual features when distinguishing between the text region and background region, it causes two limitations on the two-stage STE methods.
First, the text deletion has never utilized text information, which could be a key for understanding scene text images. 
Second, the text deletion module cannot learn from real-world examples since requiring visual supervision for text-erased backgrounds. 

In this paper, we present a novel representational learning-based STE framework, referred to as RewriteNet, which implicitly distinguishes content and style features without the explicit text deletion tasks.  
Specifically, we introduce a scene text recognition (STR) module to disentangle content features representing a series of characters from style features containing anything others such as font style, font color, text alignment, and background. In addition, to avoid mixing other visual information with content features, we detach the gradient flow from the final generation to the content features.  
With separately extracted style and content features, a generator can be trained to synthesize an image with a target text while preserving the style of the original image.
Thus, RewriteNet replaces the text deletion and conversion stages of previous work with a simple encoder in the latent space and the model can be trained in an end-to-end manner.

We also propose a self-supervised training scheme that does not require additional annotation cost and enables to exploit unlabeled real-world images. The proposed self-supervised training scheme prevents the trained model to be biased in synthetic styles and bridges the domain gap between training and test environments. As shown in Figure~\ref{fig:teaser}, our model robustly generates text-edited images where the styles of original images are well preserved. Our extensive experiments demonstrate the superiority of RewriteNet and further analysis shows that our method reliably decomposes content and style features. 
\section{Related Works}
\begin{figure*}[t]
    \centering
    \includegraphics[width=1.0\linewidth]{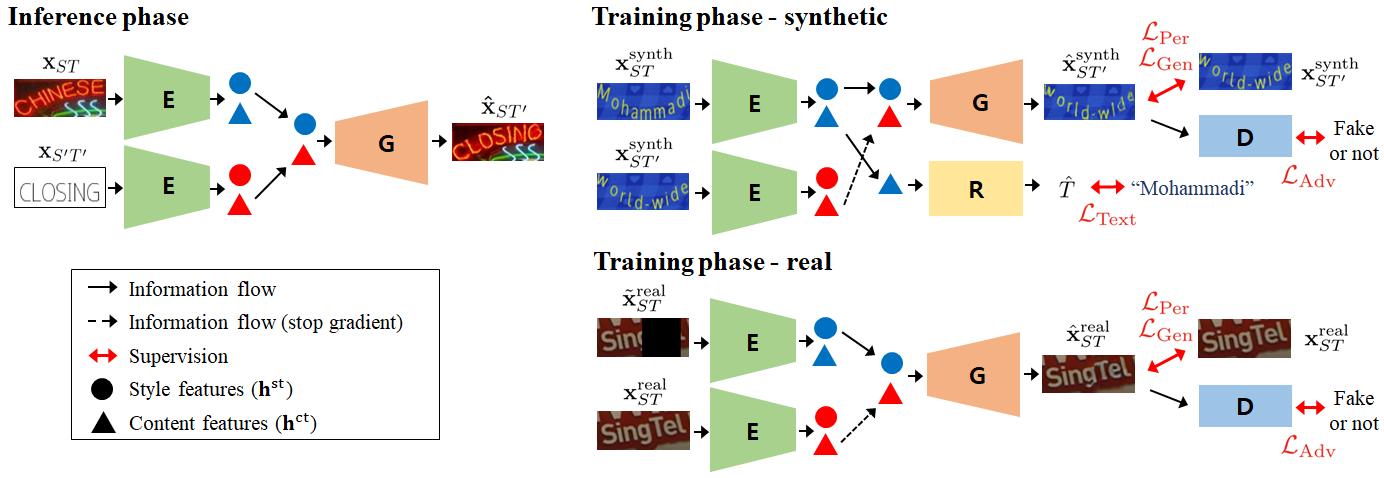}
    \vspace{-1.5em}
    \caption{
    Overview of RewriteNet during inference and training processes. RewriteNet is composed of Style-content encoder ($\mathbf{E}$), Generator ($\mathbf{G}$), Text recognizer ($\mathbf{R}$), and Discriminator ($\mathbf{D}$). In each phase, we use the same encoder to extract style and content features from two different images. The output image is generated by combining the style feature (\textcolor{iblue}{$\bullet$}) extracted from the top image (style-image) and the content feature (\textcolor{red}{$\blacktriangle$}) extracted from the bottom image (content-image).
    }
    \label{fig:overview_training}
\end{figure*}

\subsection{Scene Text Editing}

As the growth of generation model~\cite{zhu2017unpaired, zhu2017toward}, STE has been actively studied for its various applications. 
Previous STE methods mainly have proposed multiple sub-modules to extract a background and spatial text alignment, and a single fusion module to generate a text-edited image with the identified information.
Specifically, initial STE work~\cite{stefann2020} segments binary mask for each character and switches it into the desired character. 
Although it has shown that their character correction method can be applied in real-world images, 
it cannot deal with different lengths between the text in the original image and the desired text. 
Moreover, its simple rule-based segmentation module could critically affect the generation performance.

Recently, Wu \etal~\cite{wu2019etiw} and Yang \etal~\cite{swaptext2020} have proposed word-level STE methods using text background in-painting and fusion modules. 
These methods attempt to train the model to separate the text region and the background region using the text-erased image.
They could successfully conduct word-level STE, however, sometimes they fail to edit the text of the complex style images.
To deal with this problem, we exploit text information and real-world images by proposing a representational learning-based word-level STE framework.

\subsection{Image-to-Image Translation}
Image-to-image translation methods have been widely researched due to their practical usages. Isola \etal~\cite{isola2017image} have proposed paired image-to-image translation with conditional GAN to learn a mapping from the input image to the output image. To address unpaired image-to-image translation, UNIT~\cite{unit} and MUNIT~\cite{munit} assume fully and partially shared latent space, respectively.
Ma \etal~\cite{Ma2018egscit} have proposed an exemplar guided image translation method with a semantic feature mask that does not require additional labels for feature masks.
Motivated by previous works, we introduce partially shared latent space assumption and the self-supervised training scheme with STE specialized proposals.

\section{RewriteNet}
RewriteNet consists of an encoder that extracts decomposed style and content features, and a generator that generates a text image with the identified features. This section first describes how encoder and generator are utilized to generate an image of desired text, and explains how the modules are trained on synthetic and real datasets. Followed by it, architectural details are provided.

\subsection{Inference Process}

Let $\mathbf{x}_{ST}$ be the style-image with text $T$ and style $S$. When a target text $T'$ is given, our model aims to generate $\mathbf{x}_{ST'}$ whose text is switched into the target text $T'$ from $\mathbf{x}_{ST}$ while holding its style $S$. 
To achieve the goal, RewriteNet assumes two disentangled latent features, $\mathbf{h}^{\text{st}}_S$ for the style $S$ and $\mathbf{h}^{\text{ct}}_T$ for the content $T$, and conducts the content switch. 

Following the encoder and generator framework, the inference model consists of the below two modules. 
\begin{itemize}
\item \textbf{Style-content encoder} 
($\mathbf{E}$: $\mathbf{x}_{ST} \rightarrow \mathbf{h}^{\text{st}}_{S}$, $\mathbf{h}^{\text{ct}}_{T}$)
 extracts latent style feature $\mathbf{h}^{\text{st}}_{S}$ and content feature $\mathbf{h}^{\text{ct}}_{T}$ from an image $\mathbf{x}_{ST}$. 
For better descriptions, $\mathbf{E}$ will be expressed as two terms; 
 $\mathbf{E}^{\text{st}}: \mathbf{x}_{ST} \rightarrow \mathbf{h}^{\text{st}}_{S}$ and $\mathbf{E}^{\text{ct}}: \mathbf{x}_{ST} \rightarrow \mathbf{h}^{\text{ct}}_{T}$.


\item \textbf{Generator} ($\mathbf{G}$: $\mathbf{h}^{\text{st}}_{S}, \mathbf{h}^{\text{ct}}_{T} \rightarrow \hat{\mathbf{x}}_{ST}$) generates an output image of text $T$ under the style $S$. 
\end{itemize}
By switching off the latent content features, the model becomes enabled to generate a text-switched image $\hat{\mathbf{x}}_{ST'}$ as follows:
\begin{equation}
    \hat{\mathbf{x}}_{ST'} = \mathbf{G}( \mathbf{E}^{\text{st}}(\mathbf{x}_{ST}), \mathbf{E}^{\text{ct}}(\mathbf{x}_{S'T'}) ).
\end{equation}
The left of Figure~\ref{fig:overview_training} explains the inference process in a view of the information flow.
A content-image $\mathbf{x}^{\text{synth}}_{S'T'}$ is synthetically rendered with simple style $S'$ and target text $T'$.

\subsection{Training Process}

The right of Figure~\ref{fig:overview_training} shows two training processes of our model.
One is for paired synthetic images, and the other is for unpaired real-world images.

\subsubsection{Modules Utilized in Training Process}
Here, we introduce two modules only used in the training process to encourage the content-switched image generation. 
\begin{itemize}
\item \textbf{Text recognizer} ($\mathbf{R}$: $\mathbf{h}^{\text{ct}}_{T} \rightarrow T$) identifies text label from the latent content feature.
By learning content features $\mathbf{h}^{\text{ct}}$ to predict text label, the content feature can represent the text upon on the input image and is used as a content condition of $\mathbf{G}$. We should note that content feature $\mathbf{h}^{\text{ct}}$ is only trained with text label in the whole training process.
\item \textbf{Style-content discriminator} ($\mathbf{D}$: $\hat{\mathbf{x}}_{ST'}, \mathbf{x}_{ST}, \mathbf{h}^{\text{ct}}_{T'} \rightarrow [0,1]$) determines whether an input image $\hat{\mathbf{x}}_{ST'}$ is synthetically generated with a style reference $\mathbf{x}_{ST}$ and a content feature $\mathbf{h}^{\text{ct}}_{T'}$, where $\hat{x}_{ST'}$ is an output of $\mathbf{G}$. As a competitor of $\mathbf{G}$, its adversarial training improves generation quality.   
\end{itemize}
By utilizing these modules, $\mathbf{E}$ enables to identify the latent content and the $\mathbf{G}$ enables to generate high-quality images.

\subsubsection{Learning from Synthetic Data}

We train the modules to decompose style and content features by using synthetic image pairs. As shown in the top right of Figure~\ref{fig:overview_training}, synthetic image pairs share the same style but have different text contents. The content feature is learned to capture text information in an image by utilizing $\mathbf{E}^{\text{ct}}$ and $\mathbf{R}$. The encoded content feature is fed into the recognizer, and to let the recognizer predict correct labels, the encoder is trained to produce favorable content features. The style feature is learned to represent style information by allowing $\mathbf{E}^{\text{st}}$ and $\mathbf{G}$ to maintain style consistency after content switched generation.

We can obtain paired images $\{\mathbf{x}^{\text{synth}}_{ST}, \mathbf{x}^{\text{synth}}_{ST'}\}$ by feeding different texts to synthesizing engine with same rendering parameters such as background, font style, alignment, and so on.
Then, a single training set becomes $\{\mathbf{x}^{\text{synth}}_{ST}, \mathbf{x}^{\text{synth}}_{ST'}, T\}$ where $T$ is a text label. 
Therefore, $\mathbf{E}$, $\mathbf{G}$ and $\mathbf{R}$ can be trained with reconstruction and recognition losses:
\begin{gather}
    \mathcal{L}^{\text{synth}}_{\text{Gen}} = \| \mathbf{G}(\mathbf{E}^{\text{st}}(\mathbf{x}^{\text{synth}}_{ST}), \bar{\mathbf{E}}^{\text{ct}}(\mathbf{x}^{\text{synth}}_{ST'})) - \mathbf{x}^{\text{synth}}_{ST'} \|_{1}, \\
    \mathcal{L}^{\text{synth}}_{\text{Text}} = \sum_{i} \operatorname{CrossEntropy}(\mathbf{R}(\mathbf{E}^{\text{ct}}(\mathbf{x}^{\text{synth}}_{ST}))_i ,T_i),
\end{gather}
where $\bar{\mathbf{E}}^{\text{ct}}$ indicates a frozen encoder that does not get any back-propagation flow and $T_i$ represents $i$-th character of the ground truth text label. 

If we do not freeze $\mathbf{E}^{\text{ct}}$ at reconstruction loss, $\mathbf{E}$ and $\mathbf{G}$ will quickly fall into a local minimum by simply copying content-image.
Thus, we freeze the $\mathbf{E}^{\text{ct}}$  to prevent the content feature from being affected by the reconstruction loss and train $\mathbf{E}^{\text{ct}}$ only with the recognition loss.
These losses guide the model to learn the content switch, but the trained model might fail to address real-world images caused by the limitation of the synthetic styles. Here, it exists input discrepancy between training and test phases. Specifically, the input pairs share the same styles in the training phase whereas the input pairs have different styles in the inference phase.
We found that training with different styles has optimization issues, and thus, RewriteNet is trained with the same style images. We will present corresponding results in a supplemental file.  
  


\subsubsection{Learning from Real-world Data}
Synthetic data provides proper guidance for content switching, but it does not fully represent a style of real-world images. To reflect real-world styles, we propose a self-supervised training process for RewriteNet utilizing real-world data as shown in the right bottom of Figure~\ref{fig:overview_training}.

In the case of real-world images, there are no paired images that have different texts with the same style.
Moreover, it is expensive to obtain text labels of real-world images.
Therefore, we introduce conditioned denoising autoencoder loss to allow the model to learn style and content representations for unpaired real-world images.
Specifically, we cut out a region randomly selected in the width direction with length $w$ to lose some characters~\cite{cutout2017}, and then the noisy image is used as a style-image to extract the style feature from the left regions. By combining the content feature extracted from the original image, $\mathbf{G}$ fills the blank by referring to the style of the surrounding area of the blank region. 
The proposed self-supervised scheme will forbid the model to trivially autoencode style-image by using the corrupted image as style-image and enforce model to learn separated representations.
The denoising autoencoder loss is defined as:
\begin{equation} \label{dae}
\vspace{-0.1em}
    \mathcal{L}^{\text{real}}_{\text{Gen}} = \| \mathbf{G}(\mathbf{E}^{\text{st}}(\tilde{\mathbf{x}}^{\text{real}}_{ST}), \bar{\mathbf{E}}^{\text{ct}}(\mathbf{x}^{\text{real}}_{ST})) - \mathbf{x}^{\text{real}}_{ST} \|_{1},
\vspace{-0.1em}
\end{equation}
where $\tilde{\mathbf{x}}^{\text{real}}_{ST}$ indicates a noisy image corrupted from $\mathbf{x}^{\text{real}}_{ST}$. 
Here, we should note that the proposed self-supervised method does not require any text labels and paired images.

\subsubsection{Adversarial Training}

Generally, text image in the wild has high-frequency regions like complex background, diverse textures, and high contrast regions.
However, pixel-wise reconstruction loss, referred to as $\mathcal{L}_{\text{Gen}}$, has a limitation to address the high-frequency and tends to capture the low-frequencies~\cite{isola2017image}. To encourage high-frequency crispness, we apply the generative adversarial network (GAN) framework to generate realistic text images~\cite{ABP,isola2017image,cgan,lsgan}. 
Specifically, we design the $\mathbf{D}$ to represent a fake or real probability of the input image under the given conditions of its style-image and latent content. We denote $\mathbf{D}(X|X^{\text{st}}, \mathbf{h}^{\text{ct}})$ for the probability $p(X \text{ is not fake}|X^{\text{st}}, \mathbf{h}^{\text{ct}})$, where $X$ and $X^{\text{st}}$ indicate the input image and the style-image respectively. The adversarial losses are calculated as follows:   
\begin{gather}
\begin{split}
    \mathcal{L}^{\text{synth}}_{\text{Adv}} = & \log\mathbf{D}(\mathbf{x}^{\text{synth}}_{ST'}|\mathbf{x}^{\text{synth}}_{ST}, \bar{\mathbf{E}}^{\text{ct}}(\mathbf{x}^{\text{synth}}_{ST'})) \\
    & + \log\left( 1 - \mathbf{D}(\hat{\mathbf{x}}^{\text{synth}}_{ST'}|\mathbf{x}^{\text{synth}}_{ST}, \bar{\mathbf{E}}^{\text{ct}}(\mathbf{x}^{\text{synth}}_{ST'})) \right), 
\end{split} \\
\begin{split}
\mathcal{L}^{\text{real}}_{\text{Adv}} = & \log\mathbf{D}(\mathbf{x}^{\text{real}}_{ST}|\tilde{\mathbf{x}}^{\text{real}}_{ST}, \bar{\mathbf{E}}^{\text{ct}}(\mathbf{x}^{\text{real}}_{ST})) \\
    & + \log\left( 1 - \mathbf{D}(\hat{\mathbf{x}}^{\text{real}}_{ST}|\tilde{\mathbf{x}}^{\text{real}}_{ST}, \bar{\mathbf{E}}^{\text{ct}}(\mathbf{x}^{\text{real}}_{ST})) \right), 
\end{split}
\end{gather}
where $\hat{\mathbf{x}}^{\text{synth}}_{ST'}$ and $\hat{\mathbf{x}}^{\text{real}}_{ST}$ denote generated images from synthetic and real-world style-images, respectively (See Figure~\ref{fig:overview_training}). Here, it should be noted that the latent contents used as the conditions are frozen to block back-propagation flow to the $\mathbf{E}$ from the adversarial loss.

We also employ feature matching loss that stabilizes the training of various GAN models \cite{salimans16improved,wang18highres,liu19funit}. Specifically, we extract intermediate feature maps of the $\mathbf{D}$ and minimize the distance between generated and target samples:
\begin{gather}
    \mathcal{L}^{\text{synth}}_{\text{Per}} = \sum_l \frac{1}{M_l} \| \phi_l(\mathbf{x}^{\text{synth}}_{ST'}) - \phi_l(\hat{\mathbf{x}}^{\text{synth}}_{ST'}) \|_1, \\
    \mathcal{L}^{\text{real}}_{\text{Per}} = \sum_l \frac{1}{M_l} \| \phi_l(\mathbf{x}^{\text{real}}_{ST}) - \phi_l(\hat{\mathbf{x}}^{\text{real}}_{ST}) \|_1,
\end{gather}
where $\phi_l$ and $M_l$ are the output feature map and its size of the $l$-th layer. For each loss, the same conditions are used to calculate the activation maps $\{ \mathbf{x}^{\text{synth}}_{ST}, \bar{\mathbf{E}}^{\text{ct}}(\mathbf{x}^{\text{synth}}_{ST'})\}$ for $\mathcal{L}^{\text{synth}}_{\text{Per}}$ and $\{ \tilde{\mathbf{x}}^{\text{real}}_{ST}, \bar{\mathbf{E}}^{\text{ct}}(\mathbf{x}^{\text{real}}_{ST})\}$ for $\mathcal{L}^{\text{real}}_{\text{Per}}$. 
Feature matching losses could facilitate $\mathbf{G}$ to match multi-scale statistics with target samples \cite{wang18highres}, thus beneficial for overall sample qualities.

\subsubsection{Final Loss Term}

The final losses are formalized as follows:
\begin{gather}
\begin{split}
    \mathcal{L} \; = \argmin_{\mathbf{E}, \mathbf{G}, \mathbf{R}} &  \Big( \mathcal{L}^{\text{synth}}_{\text{Gen}} +  \mathcal{L}^{\text{synth}}_{\text{Text}} +  \mathcal{L}^{\text{synth}}_{\text{Per}} \\
    & \; + \left(\mathcal{L}^{\text{real}}_{\text{Gen}} +  \mathcal{L}^{\text{real}}_{\text{Per}} \right) \\
    & \; + \lambda \; \argmax_{\mathbf{D}} \left( \mathcal{L}^{\text{synth}}_{\text{Adv}} +  \mathcal{L}^{\text{real}}_{\text{Adv}} \right) \Big),
\end{split}
\end{gather}
where $\lambda$ is intensity balancing the losses. 

\subsection{Architectural Details}
\paragraph{Style-content Encoder}
The style-content encoder follows \emph{partially shared latent space assumption} as in MUNIT \cite{huang2018multimodal}, where an image $\mathbf{x}_{ST}$ is composed of its latent style feature $\mathbf{h}^{\text{st}}_{S}$ and content feature $\mathbf{h}^{\text{ct}}_{T}$. 
The network is based on a ResNet~\cite{ResNet} similar to the feature extractor used in \cite{FAN}. 
In addition, we apply bidirectional LSTM~\cite{LSTM} layers upon the content features to alleviate spatial dependencies from the input image. 

\paragraph{Text recognizer}
Text Recognizer estimates a sequence of characters in an image and it has an important role to distinguish contents from styles. It consists of an LSTM decoder with an attention mechanism~\cite{baek2019STRcomparisons} from the identified content features. Since the text labels are required to train this module, we only train the module with the synthetic dataset.

\paragraph{Generator}
Given the latent style and content features as an input, the generator outputs an image with a given style and content. The generator network is similar to the decoder used in the Unet~\cite{ronneberger2015u} architecture. The style features in multiple $\mathbf{E^{\text{st}}}$ layers are fed into the generator using short-connections. The network design is inspired by the Pix2Pix~\cite{isola2017image} model.

\paragraph{Style-content discriminator}
The style-content discriminator determines whether an image is fake or not. 
The network architecture follows PatchGAN~\cite{isola2017image,shrivastava2017learning}.

\section{Experiments}
\begin{figure*}[t!]
    \centering
    \includegraphics[width=0.9\linewidth]{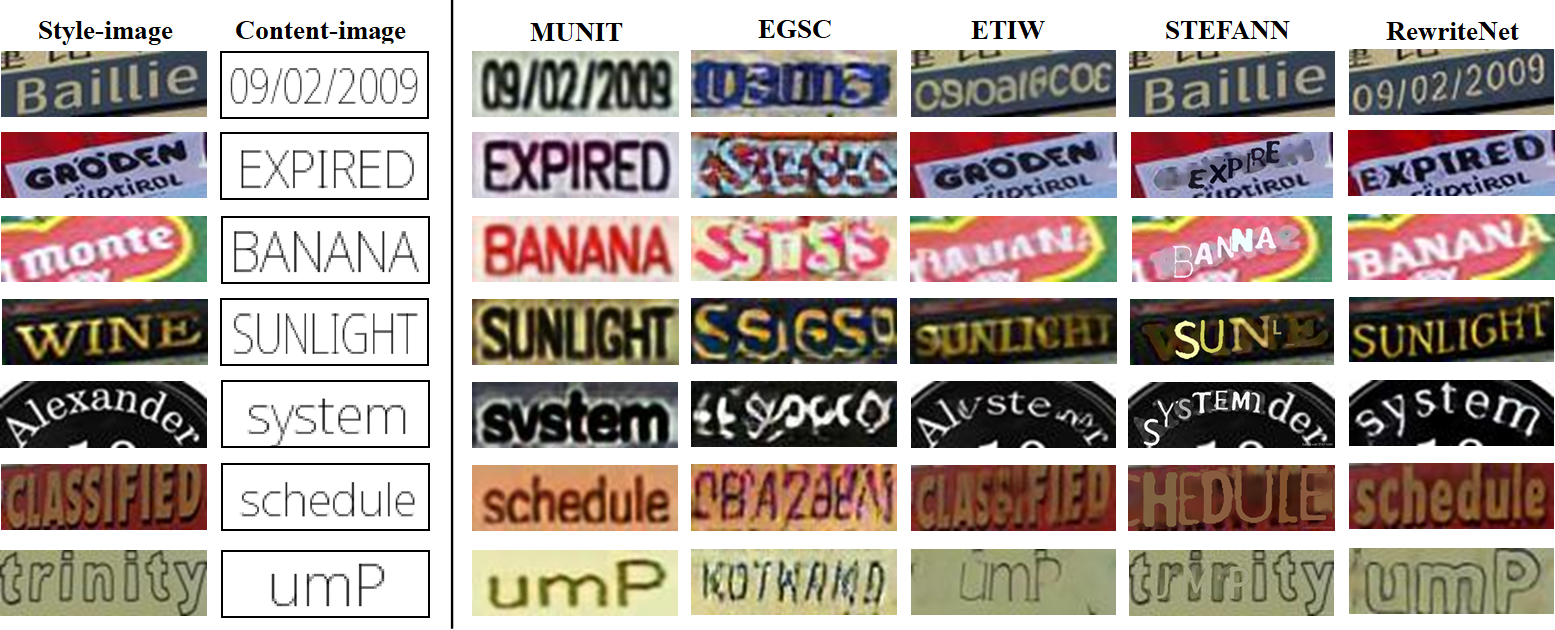}
    \caption{Visual comparisons on text-editing performance. Target texts are ``09/02/2009'', ``EXPIRED'', ``BANANA'', ``SUNLIGHT'', ``system'', ``schedule'' and ``umP'', respectively.} 
    \label{fig:gen_comparision}
\end{figure*}

\subsection{Datasets and Implementation Details}

\subsubsection{Synthetic Data for Training}
Since RewriteNet requires paired synthetic datasets for supervised-learning, we generate 8M synthetic images with public synthesizing engine\footnote{https://github.com/clovaai/synthtiger} that is based on MJSynth~\cite{Jaderberg16mjsynth} and SynthText~\cite{Gupta16synthtext}.
Specifically, we compose paired data $\{\mathbf{x}^{\text{synth}}_{ST}, \mathbf{x}^{\text{synth}}_{ST'}\}$ with same rendering parameters $S$ such as font styles, background textures, shape for the text alignments (rotation, perspective, curve), and artificial blur noises except only for input texts ($T, T'$).   
The employed texts are the union of MJSynth and SynthText corpus and the paired synthetic dataset will be publicly available.


\subsubsection{Real Data for Training and Evaluation}
We combine multiple benchmark training datasets such as IIIT~\cite{IIIT5K}, IC13~\cite{IC13}, IC15~\cite{IC15}, and COCO~\cite{COCO}. The total number of training images is 59,856. Although these datasets contain ground-truth text labels, our model does not employ the text labels that are expensive in a practical scenario. 
For evaluation, we use the test a split of each public dataset such as IIIT~\cite{IIIT5K}, IC03~\cite{IC03}, IC13~\cite{IC13}, IC15~\cite{IC15}, SVT~\cite{SVT}, SVTP~\cite{SVTP} and CT80~\cite{CUTE80} where the total number of test images is 8,536. 

\subsubsection{Implementation Details}
We rescale the input image to 32 $\times$ 128 and empirically set $w$ as 42, which is the proper length to cut out some characters and capture style information for the self-supervised training.
To balance the multiple loss terms, we empirically set $\lambda=0.1$. The model is optimized by Adam optimizer~\cite{adam} with $\beta_1=0.9$ and $\beta_2=0.999$. A cyclic learning rate~\cite{smith2017cyclical} is applied with an initial learning rate of 1e-4 and an maximum iteration number of 300K. The batch size is 192 including 144 for synthetic data and 48 for real-world data. The total training takes 7 days using two Tesla V100s.
At the inference phase, $\mathbf{x}^{\text{synth}}_{S'T'}$ is generated by $ImageDraw$ function from $PIL$ package.

\subsection{Comparison on Generation Performance} \label{sec:comparison_gp}
We compare our model to four models:
MUNIT~\cite{huang2018multimodal}, EGSC~\cite{Ma2018egscit}, ETIW~\cite{wu2019etiw}, and STEFANN~\cite{stefann2020}. 
Although MUNIT and EGSC are not specifically designed for STE task, we train the model targeted for STE task and make a comparison with our model to validate the STE performance of the representative image translation models\footnote{We use the official codes: MUNIT(\url{https://github.com/NVlabs/MUNIT}) and EGSC(\url{https://github.com/charliememory/EGSC-IT})}.
ETIW is the exact comparison method for RewriteNet and its results are achieved from  re-implementation\footnote{https://github.com/youdao-ai/SRNet}.
STEFANN is designed for the character-wise correction method that requires manual text region selection, so test environments are different from other methods. We try to achieve high-quality results for STEFANN by testing multiple times with its official code\footnote{https://github.com/prasunroy/stefann}.

\begin{table}[t]
\begin{center}
\tabcolsep=0.15cm
\caption{Quantitative comparison between STE methods:
``Accuracy'' represents the content-switch performance (higher is better) and ``FID'' shows style consistency (lower is better). The bold indicates the best performance.}
\begin{tabular}{lcc}
\toprule 
\textbf{Models} & \textbf{Accuracy} ($\uparrow$) &  \textbf{FID} ($\downarrow$)  \\
\midrule \midrule
MUNIT~\cite{huang2018multimodal} & 80.75 &	65.7 \\
EGSC~\cite{Ma2018egscit} & 0.04 &	43.3 \\
ETIW~\cite{wu2019etiw} & 31.16  &	\textbf{13.7} \\
\midrule
Ours &  \textbf{90.30} &  16.7  \\
\bottomrule
\end{tabular}
\label{table:comparison_generations}
\end{center}
\end{table}

\begin{figure*}[t]
    \centering
    \includegraphics[width=0.85\linewidth]{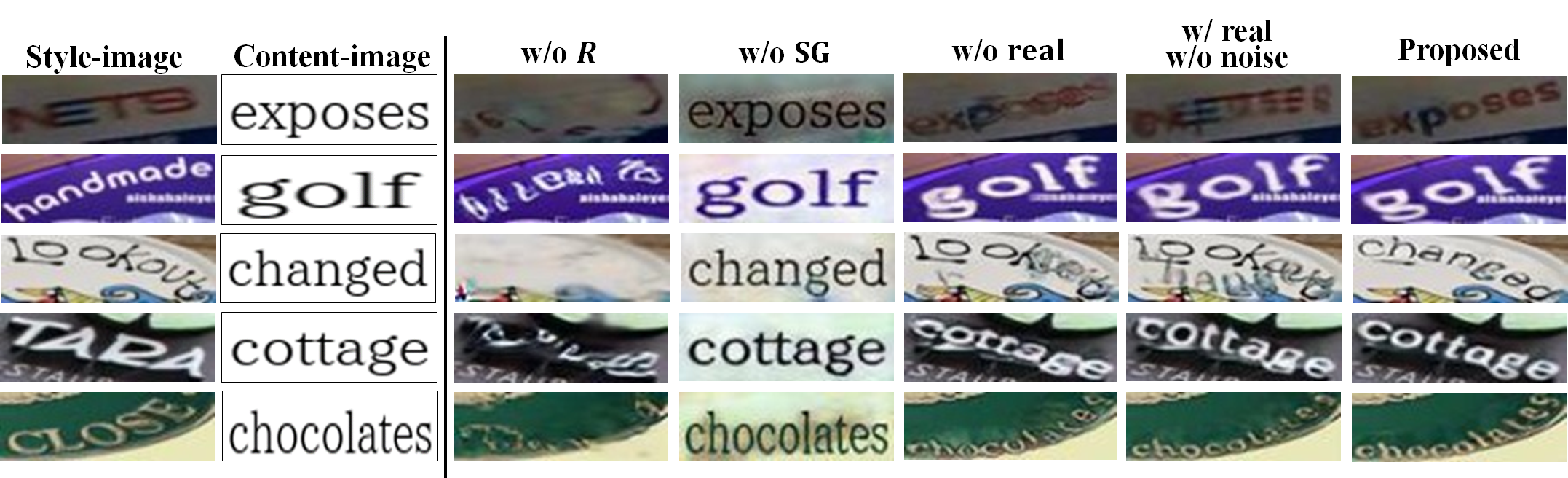}
    \caption{Generated images from RewriteNet trained with ablated training processes. ``w/o SG'' indicates the model is trained without stop gradient. 
    Target texts are ``exposes'', ``golf'', ``changed'', ``cottage'' and ``chocolates'', respectively.}
    \label{fig:ablative_dae}
\end{figure*}

In the quantitative comparison, we employ two measurements: recognition accuracy on generated images utilizing a pre-trained STR model\footnote{https://github.com/clovaai/deep-text-recognition-benchmark}, and Fréchet Inception Distance (FID)~\cite{heusel2017gans}. The recognition accuracy measures whether the generated images truly contain switched contents or not. The FID represents style consistency between a style-image and a generated image. 
Here, we would note that measurements between text switching performance and style preserving performance have a trade-off.
It is because the best performance on FID is achieved when the output is the same as the input.
Thus, balanced quantitative performance and visual results should be considered to compare the model performance.
Quantitative performance on STEFANN is not evaluated, because it requires manual region selection for each image and it considerably takes a long time.

Table~\ref{table:comparison_generations} presents the quantitative comparison results. 
The naive application of MUNIT tends not to maintain original styles, which can be confirmed in its high FID score. 
We observe that the naive application of EGSC would be inappropriate for STE.
ETIW shows the best performance on FID, however, it achieves comparably lower accuracy.
These results indicate ETIW often fails to convert the content and simply copies style-image 
where the examples are shown in Figure~\ref{fig:gen_comparision} (2nd and 6th rows). 
The proposed model achieves the best accuracy and also shows comparable performance on FID. 

The visual comparisons are presented in Figure~\ref{fig:gen_comparision}. MUNIT looks failed to preserve the style of style-image and ETIW tends to simply copy style-image for challenging style without content switching. STEFANN also cannot robustly edit texts when the lengths of texts are different and backgrounds are complex.
In contrast, the proposed methods show promising results on multiple examples compared to other methods. 
More visual results are presented in Figure~\ref{fig:teaser} where multi texts also can be edited by employing text region detector~\cite{baek2019character}. More visual results will be presented in supplemental file.

\begin{table}[t]
\begin{center}
\caption{Ablation study about the training processes.}
\tabcolsep=0.1cm
\begin{tabular}{lcc}
\toprule
\textbf{Models} & \textbf{Accuracy} ($\uparrow$) &  \textbf{FID} ($\downarrow$)  \\
\midrule\midrule
Proposed & 90.30 &  16.7 \\
\midrule
w/o $\mathbf{R}$ & 0.54  &  114.9 \\
w/o Stop Gradient  & 97.22  &  89.9 \\
\midrule
w/o real & 89.00 &	18.7 \\ 
w/ real w/o noise & 82.97 &		20.9 \\ 
\bottomrule
\end{tabular}
\label{table:ablative_number}
\end{center}
\end{table}

\subsection{Ablation Study}
In RewriteNet, the feature decomposition is conducted by the use of a recognizer and stop gradient. 
Here, we describe the effectiveness of employing the recognizer and stop gradient with ablated training processes: a model without the recognizer (w/o $\mathbf{R}$) and a model without the stop gradient (w/o Stop Gradient).
Table~\ref{table:ablative_number} and Figure~\ref{fig:ablative_dae} show the comparison results. 
We observe that RewriteNet cannot be trained without $\mathbf{R}$ where the performance of text switching and style preservation is dramatically degraded.  
``w/o Stop Gradient'' achieves higher accuracy than ours, however, the performance of FID is much worse. The visual results also present the necessity of $\mathbf{R}$ and stop gradient. Specifically, ``w/o Stop Gradient'' simply writes the desired texts without preserving styles, which is quantitatively shown.

Moreover, we also validate the use of real data and self-supervised learning scheme with ablated training processes: a model without the training branch utilizing real-world data (w/o real) and a model feeding an original real-world image into the training branch instead of its noisy variant (w/ real w/o noise).
As presented in Table~\ref{table:ablative_number}, ``w/o real'' and  ``w/ real w/o noise'' achieve lower quantitative performance than the proposed method and visual results in Figure~\ref{fig:ablative_dae} also show that the use of real data with noises generates clearer visual results when the texts are irregularly shaped and background are complex.

\begin{figure}
  \centering
  \begin{subfigure}{0.99\linewidth}
    \includegraphics[width=1.0\linewidth]{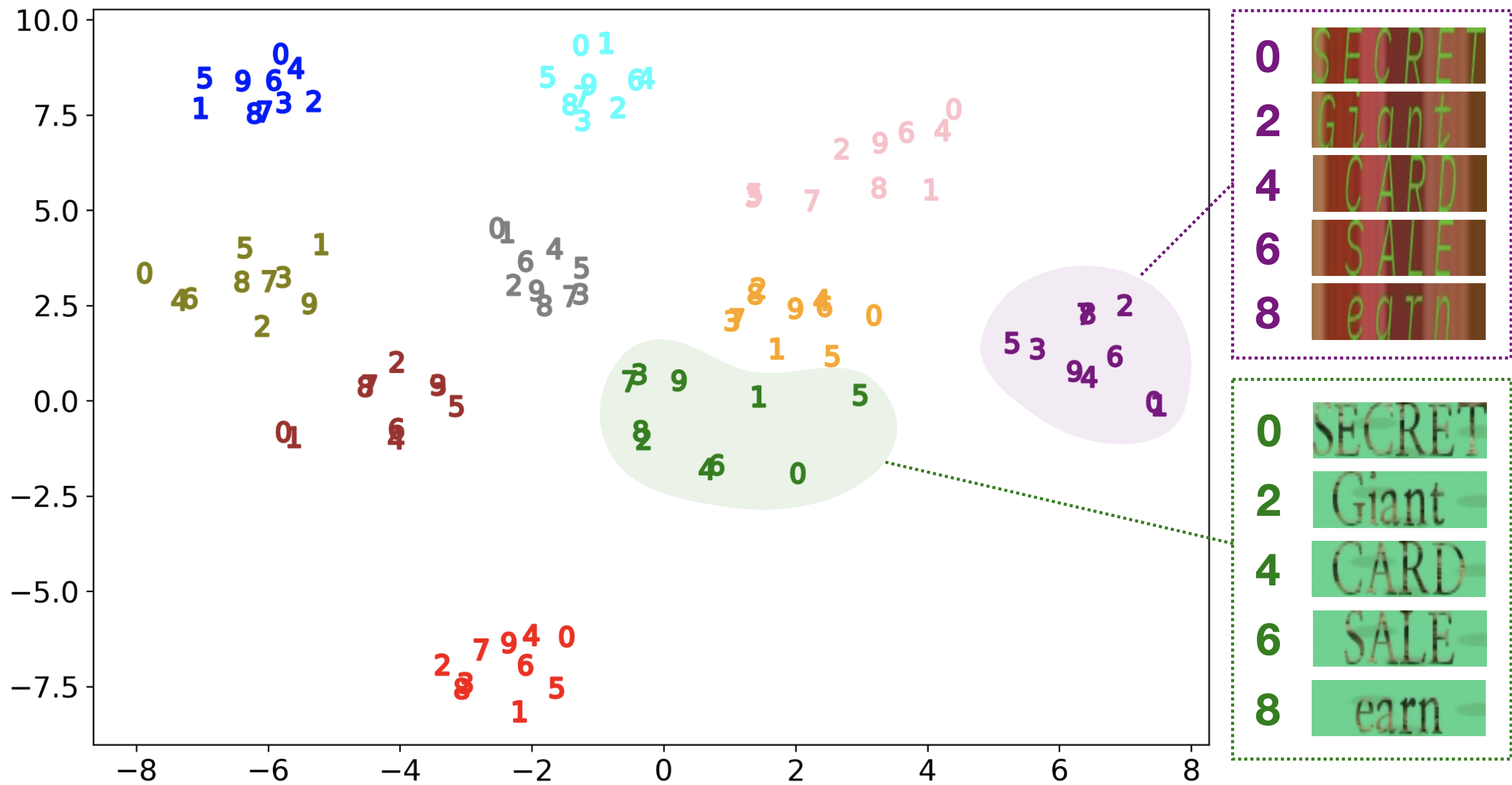}
    \caption{T-SNE visualization of style features.}
    \label{fig:tsne_style}
  \end{subfigure}
  \hfill
   \begin{subfigure}{0.99\linewidth}
    \includegraphics[width=1.0\linewidth]{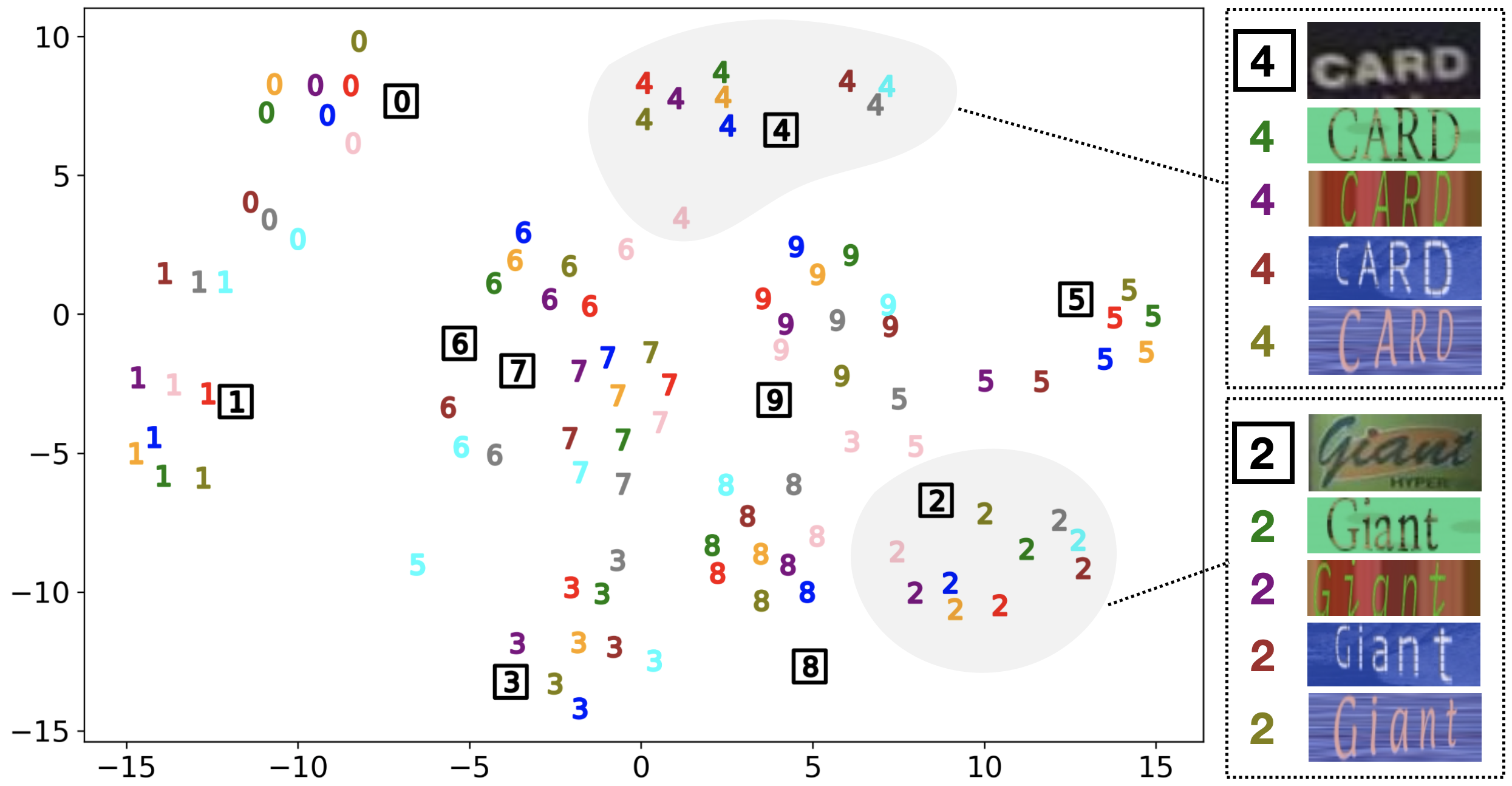}
    \caption{T-SNE visualization of content features.}
    \label{fig:tsne_content}
  \end{subfigure}
  \caption{Visualizations of decomposed style and content features. The colored numbers indicate synthetic examples including 10 contents (numbers) with 10 styles (colors) and the boxed numbers represent real-world examples. The same styles and contents are placed closely in the corresponding feature spaces. The content features of the real-world examples are involved in the clusters holding the same text.}
\label{fig:tsne}
\end{figure}
\subsection{Discussions}
\subsubsection{Content and Style Decomposition} \label{sec:ablation1}

To validate whether our model successfully separates the content feature from the style feature, we investigate style and content features of 10$\times$10 synthetic images (10 contents and 10 styles). Figure~\ref{fig:tsne} shows the T-SNE visualizations. As can be seen, the same styles (represented with colors) are plotted closely in the style feature space and the same contents (represented with numbers) are grouped in the content feature space. In addition, we also explore the content features of real-world examples, which have the same contents as the synthetic samples, and observe that they are involved in the corresponding content clusters. 

\begin{figure}[t!]
    \centering
    \includegraphics[width=0.95\linewidth]{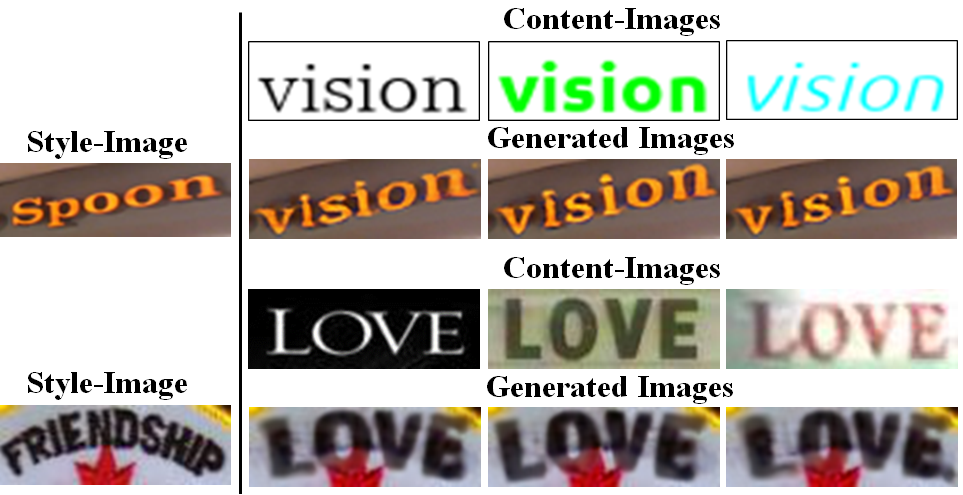}
    \caption{Generated images when given diverse content-images that have different styles with the same content}
    \label{fig:ablative_invariance}
    \vspace{1.0em}
    \includegraphics[width=1\linewidth]{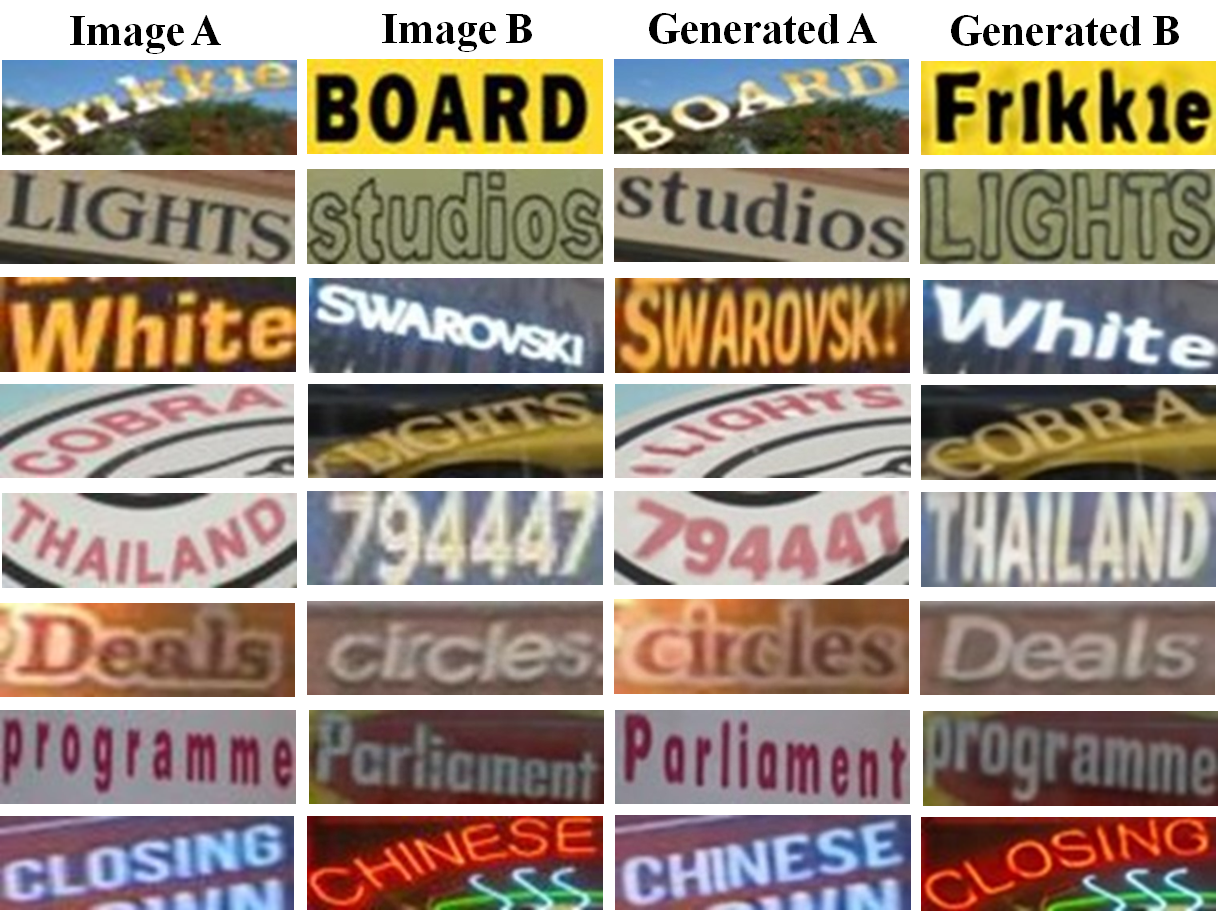}
    \caption{Text switched images when the content and style images are real scene images. ``Image A'' and ``Image B'' are the input images for generations. ``Generation A'' brings style and content from  ``Image A'' and ``Image B'', respectively. Similarly, ``Generation B'' brings style and content from ``Image B'' and ``Image A'', respectively.}
    \label{fig:real_results}
\end{figure}

We also show that the style of content-image does not affect the style of the generated image to validate feature decomposition. 
We feed various images for content-images that have different styles with the same content and observe whether the generated results are affected.
As shown in Figure~\ref{fig:ablative_invariance}, the generated results are quite stable to the change of content-images.
Furthermore, direct text switching between real images could be conducted by switching the content features. As shown in Figure~\ref{fig:real_results}, ``Generated A'' and ``Generated B'' can be achieved with great visual quality when style and content images are real images. 
These results validate that our model well separates the content feature from the style feature of input images.

\subsubsection{Robustness for Text Lengths of Contents}
\begin{figure}[t]
    \centering
    \includegraphics[width=1.0\linewidth]{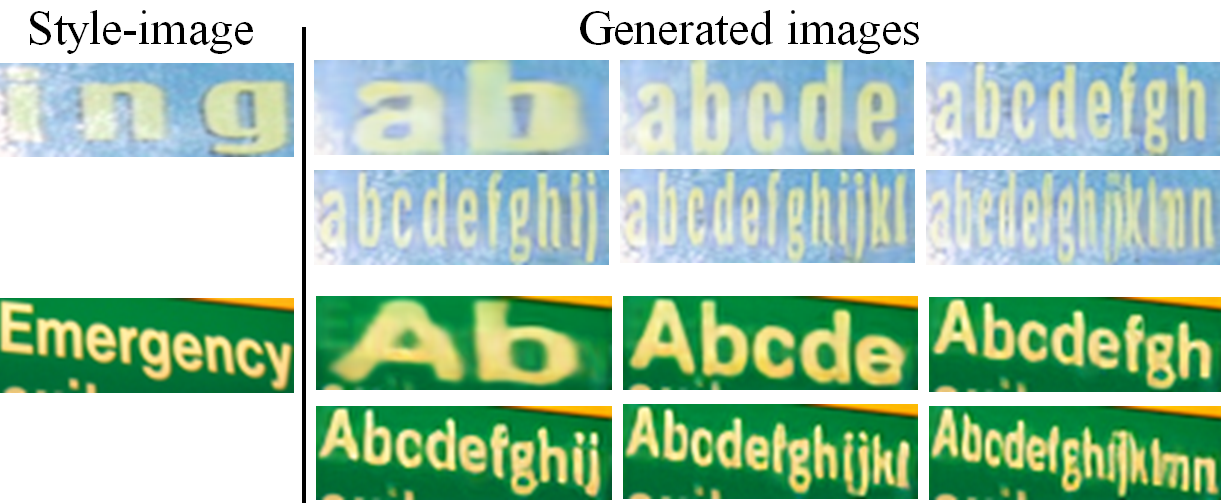}
    \caption{Generated images according to the change of lengths of the desired text.}
    \label{fig:different_lengths}
    \vspace{-0.5em}
\end{figure}

To show the robustness of text editing for different text lengths between desired text and style-image, we present more generation examples when the lengths of the desired texts are extremely different from the text of style-image.
As can be seen in Figure~\ref{fig:different_lengths}, RewriteNet can edit different length texts robustly. Specifically, RewriteNet generates great quality outputs (1st example) when converting the 3 characters (ing) to 14  characters (abcdefghijklmn).
Interestingly, we observe that the model can properly adjust the height, width, and spacing of characters as the number of characters changed.


\begin{table}[t]
\tabcolsep=0.09cm
\centering
\caption{STR accuracy over three benchmark test datasets depending on the training data.
``Synth'' indicates font-rendered data from MJSynth and SynthText.
``MUNIT'', ``ETIW'', and ``Ours'' represent fully generated data from unlabeled real images using MUNIT, ETIW, and RewriteNet, respectively. 
}
\begin{tabular}{cc|l|ccc} 
\toprule
& Model & Train Data & IC15 & CUTE80 & SVTP \\
\midrule
\midrule
&TRBA & Synth &  78.0 & 76.7 & 79.5 \\
\midrule
&TRBA & Synth+MUNIT &  62.0 \redp{$\downarrow$} &	57.3 \redp{$\downarrow$}&	66.2 \redp{$\downarrow$}\\
&TRBA & Synth+ETIW & 64.0 \redp{$\downarrow$} & 62.8 \redp{$\downarrow$} & 61.1 \redp{$\downarrow$} \\
&TRBA & Synth+Ours & \textbf{79.6} \greenp{$\uparrow$} & \textbf{84.4} \greenp{$\uparrow$} & \textbf{81.6} \greenp{$\uparrow$} \\
\bottomrule
\end{tabular}

\label{table:more_trainingset}
\end{table}

\subsubsection{Learning from Text Edited Images}

It is well-known that an accurate training set leads to better performances. 
To evaluate the reliability of the text-edited images, we utilize the generated images for training STR models and investigate the performance gains.
We train TRBA~\cite{baek2019STRcomparisons}, a popular STR baseline, with the generated examples and rule-based synthetic images~\cite{Jaderberg16mjsynth,Gupta16synthtext}. 
For the generation, four benchmark training datasets such as IIIT~\cite{IIIT5K}, IC13~\cite{IC13}, IC15~\cite{IC15}, and COCO~\cite{COCO} are employed as the style-images and the total amount of generated images is 1M. 
Here, we would note that STE methods do not require additional text labels for generating samples and training iterations are same for all comparisons even if the training data increases.  

In Table~\ref{table:more_trainingset}, we observe other comparison methods including STE method are harmful to train STR model. 
These performance degradation might result from noise labels where models cannot reliably edit texts and make a mismatch between images and labels.
On the other hand, the proposed RewriteNet contributes to the performance improvement all benchmarks. 
The results prove that RewriteNet provides more accurate and reliable examples enough to be used for training text recognition models. We will present more results according to the change of STR models~\cite{CRNN, shi2016robust} in supplemental file.
\subsection{Limitations}
Most of STEs including RewriteNet convert text patch-wisely, which inevitably generates unnatural boundaries around the edited patches in the entire scene images.  We expect that an end-to-end scheme incorporating scene text detection and STE would relieve this problem.


\section{Conclusions}
This paper proposes RewriteNet which edits text in a scene image via implicit decomposition of style and content features. The novel feature decomposition methods through STR network successfully distinguish content and style features, and their combinations are used to generate text-edited images. Thanks to the simplified pipeline, RewriteNet can utilize unlabeled real-world images with the proposed cutout strategy to reduce a gap between synthetic and real-world domains. Compared to previous STE and image translation methods, the outputs generated by RewriteNet achieve better generation quality. Further analysis demonstrates the robustness of RewriteNet on multiple content types and the reliability on the text contents in the generated images. 

{\small
\bibliographystyle{ieee_fullname}
\bibliography{egbib}
}
\newpage
\begin{appendix}

\section{Training and Inference Strategies}
The input images for training and inference phases are slightly different in RewriteNet.
Specifically, in the synthetic training phase, the inputs \{$\mathbf{x}_{ST}$,$\mathbf{x}_{ST'}$\} of RewriteNet have same styles. On the other, in the inference phase, the inputs \{$\mathbf{x}_{ST}$,$\mathbf{x}_{S'T'}$\} of RewriteNet have different styles. This strategy is determined empirically.
In the early design choices, we found that the use of different styles in training phases 
could not achieve sufficient performance.
The training with different styles increases training difficulties and results in unstable convergence in adversarial training. 
Consequently, it generates undesirable artifacts on the characters and achieves lower text-switching performance than the same styles (ours), which can be seen in Table~\ref{table:different_pairs}.
Thus, we utilize the same styled image pairs with ``stop gradient'' that can
prevent simple auto-encoding and ensures style disentanglement.

\section{Experiments}
\subsection{Generated Examples from Full Scene Text Images}
\begin{figure*}[t]
    \centering
    \includegraphics[width=1\linewidth]{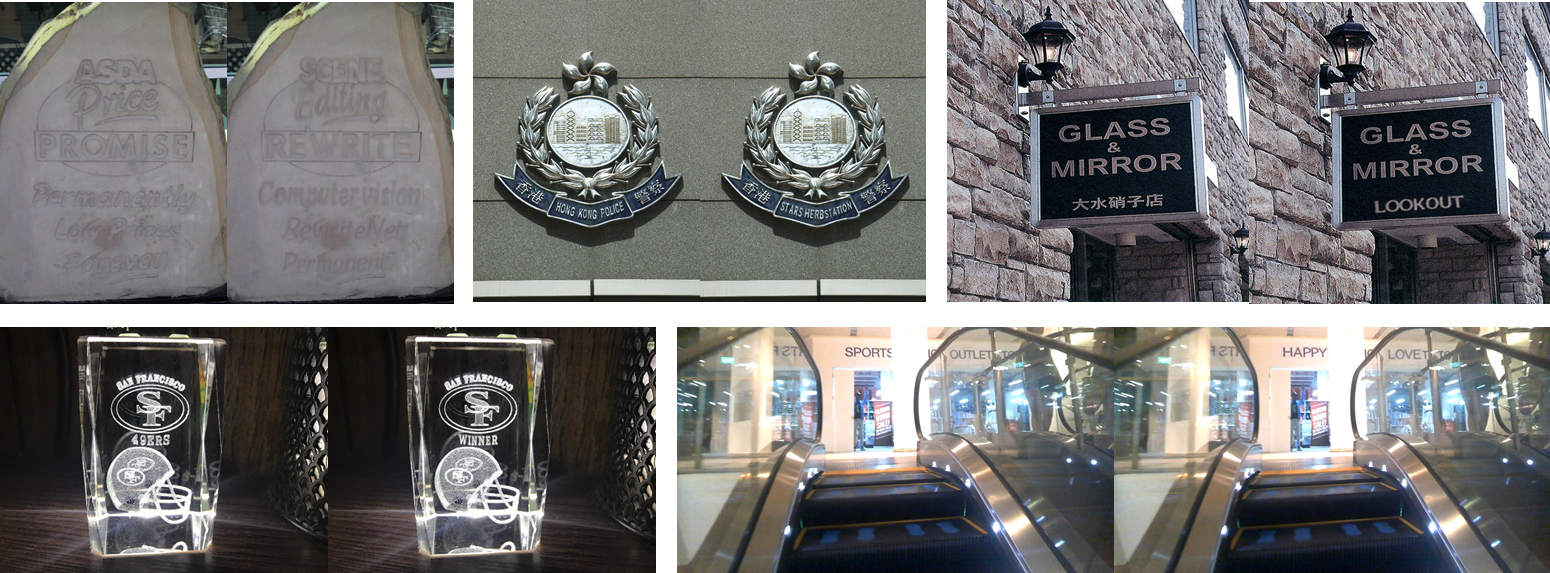}
    \caption{Full scene pairs of original (left) and text edited (right) images.}
    \label{fig:full_image}
\end{figure*}

We present more full scene generated examples by employing text detection method~\cite{baek2019character}. As shown in Figure~\ref{fig:full_image}, RewriteNet can successfully edit full scene images.  

\subsection{Comparison on Generation Performance}
The recent scene text edit method is SwapText~\cite{swaptext2020}, however, we cannot achieve code.
Thus, we simulate test environments of SwapText and we indirectly compare RewriteNet with SwapText. 
To compare content switching performance, we train the same recognizer model (CRNN) with real datasets~\cite{IC13,IC15,SVT,IIIT5K} and evaluate the recognition accuracy on real (original) and text switched images. 
In Table~\ref{table:comp_swapText}, ``Real'' achieves similar performance with ``Real*'' on SVTP and IC15, which shows that the CRNN is similarly reproduced.
Then, the generated images from RewriteNet are evaluated with CRNN and the accuracy is also reported in Table~\ref{table:comp_swapText}.
It shows that Ours achieves much higher accuracy than SwapText on all real datasets, which indicates that the proposed RewriteNet shows better text switching performance.   
 
\subsection{Ablation Study:Consistency Loss}
\begin{figure}[t]
    \centering
    \includegraphics[width=0.85\linewidth]{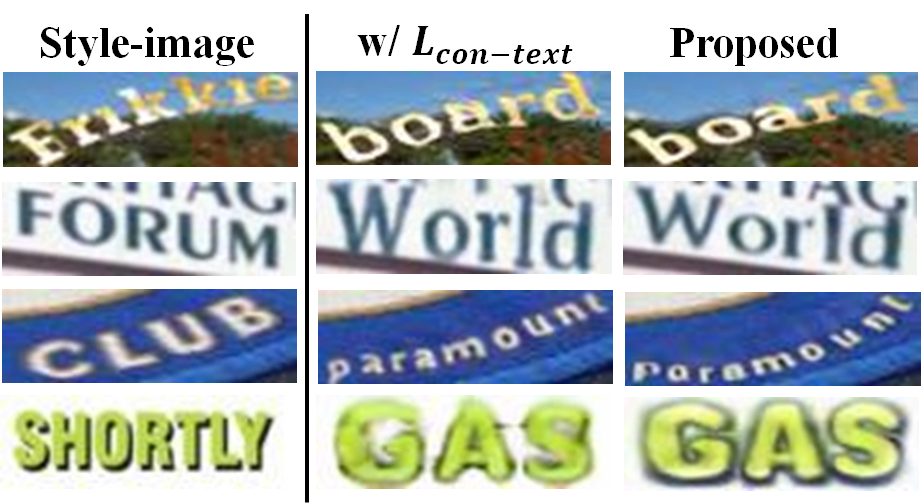}
    \caption{Visual comparisons between w/$\mathcal{L}_{\text{con-text}}$ and the proposed models.
    }
    \label{fig:ablation_consistency}
\end{figure}

Consistency loss is widely adopted in generation tasks~\cite{zhu2017unpaired,munit,cycada,stargan}, because it can improve performance by regularizing the generator. 
Following the previous works, we train RewriteNet with additional consistency loss as follows:
\begin{gather}
    \mathcal{L}^{\text{synth}}_{\text{con-text}} = \sum_{i} \operatorname{CrossEntropy}(\mathbf{R}(\mathbf{E}^{\text{ct}}(\mathbf{\hat{x}}^{\text{synth}}_{ST}))_i ,T_i),
\end{gather}
where ${\hat{x}}^{\text{synth}}_{ST}$ denotes the generated image with the style $S$ and content $T$, and $T_i$ represents $i$-th character of the ground truth text label. This loss re-enforces the generated image to have desired text and it can only be applied on the synthetic data due to the requirement of the text label. 
The quantitative performance is reported in Table~\ref{table:ablation_consistency}. It achieves a higher accuracy than the proposed, however, visual results are worse than the proposed method
as can be seen in Figure~\ref{fig:ablation_consistency}, ``w/ $\mathcal{L}_{\text{con-text}}$'' 
sometimes erases text that is out of interest in the background (part of characters are erased in the top right region), for enhancing the recognition accuracy on the generated image.
Moreover, it fails to preserve font information when the text shapes and textures are complex.
From these observations, we conclude that consistency loss could not be effective to our RewriteNet and we have omitted consistency loss in the final loss term.

\begin{table}[t]
\begin{center}
\tabcolsep=0.15cm
\begin{tabular}{lcc}
\toprule 
\textbf{Models} & \textbf{Accuracy} ($\uparrow$) &  \textbf{FID} ($\downarrow$)  \\
\midrule \midrule
Different styles &  74.97 &  15.2  \\
Same styles (Ours) &  89.00 & 18.7   \\
\bottomrule
\end{tabular}
\caption{Quantitative comparison between the use of same (Ours) and different style images as the training dataset:
``Accuracy'' represents the content-switch performance (higher is better) and ``FID'' shows style consistency (lower is better). Two models are trained only with synthetic data.}
\label{table:different_pairs}
\end{center}
\end{table}

\begin{table}[t]
\begin{center}
\begin{tabular}{lccc}
\toprule 
\textbf{Datasets} & \textbf{SVTP}& \textbf{IC13} &  \textbf{IC15} \\
\midrule \midrule
Real*  & 54.3 & 68.0 & 55.2 \\
SwapText*  & 54.1 & 68.3 & 54.9 \\
\midrule
Real  & 53.0 & 74.2 & 55.3 \\
Ours & 66.0 & 79.7 & 74.6  \\
\bottomrule
\end{tabular}
\end{center}
\caption{Text recognition accuracy on real and generated images. ``Real'' indicates that the input images of the recognizer are the original images. On the other hand, the input images of SwapText and Ours are generated (text switched) images. 
Since the codes of SwapText is not available, we bring the performances of Real and SwapText from paper, and these performances are denoted as *.
}
\label{table:comp_swapText}
\end{table}

\begin{table}[t]
\begin{center}
\tabcolsep=0.15cm

\begin{tabular}{lcc}
\toprule 
\textbf{Models} & \textbf{Accuracy} ($\uparrow$)  & \textbf{FID} ($\downarrow$)  \\
\midrule \midrule
w/$\mathcal{L}_{\text{con-text}}$  & 94.34 &  16.7 \\ 
Ours & 90.30 &  16.7  \\
\bottomrule
\end{tabular}
\end{center}
\caption{Validations of consistency loss. ``Ours'' indicates w/o $\mathcal{L}_{\text{con-text}}$.
}
\label{table:ablation_consistency}
\end{table}

\begin{figure*}[t!]
    

    \begin{subfigure}{1.0\textwidth}
         \centering
         \includegraphics[width=\textwidth]{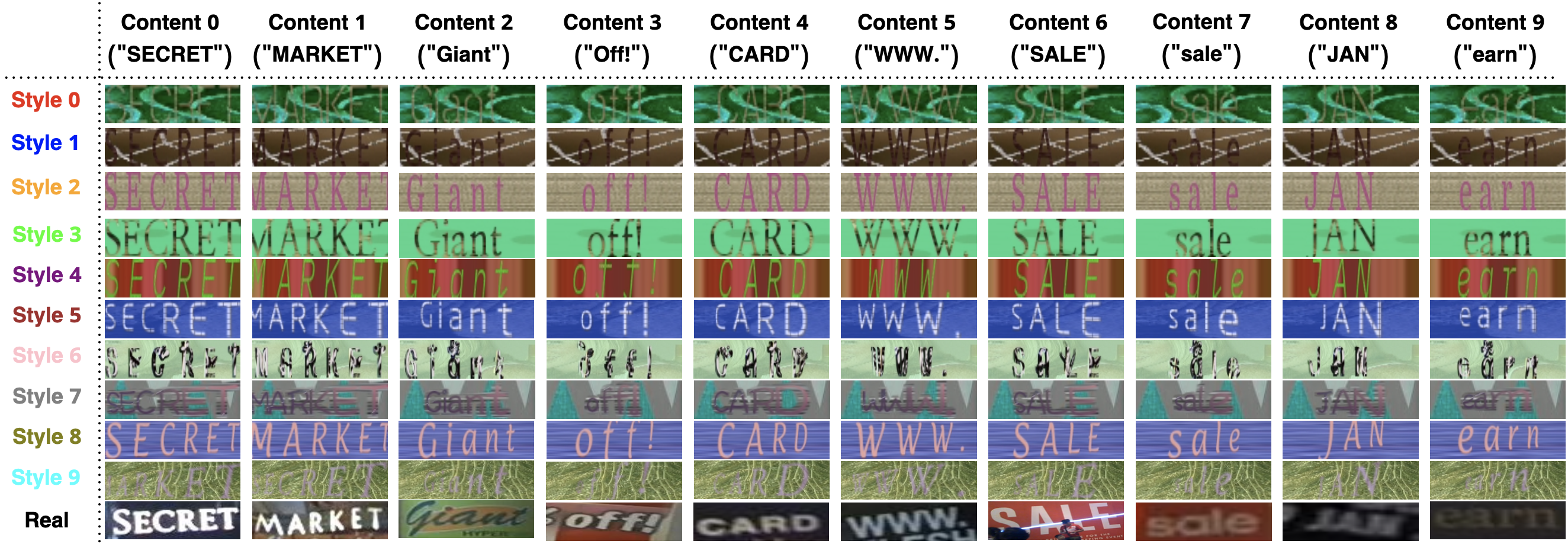}
         \caption{Visualizations of $10\times10$ synthesized images and 10 real images.}
     \end{subfigure}
     
     \begin{subfigure}{0.48\textwidth}
         \centering
         \includegraphics[width=\textwidth]{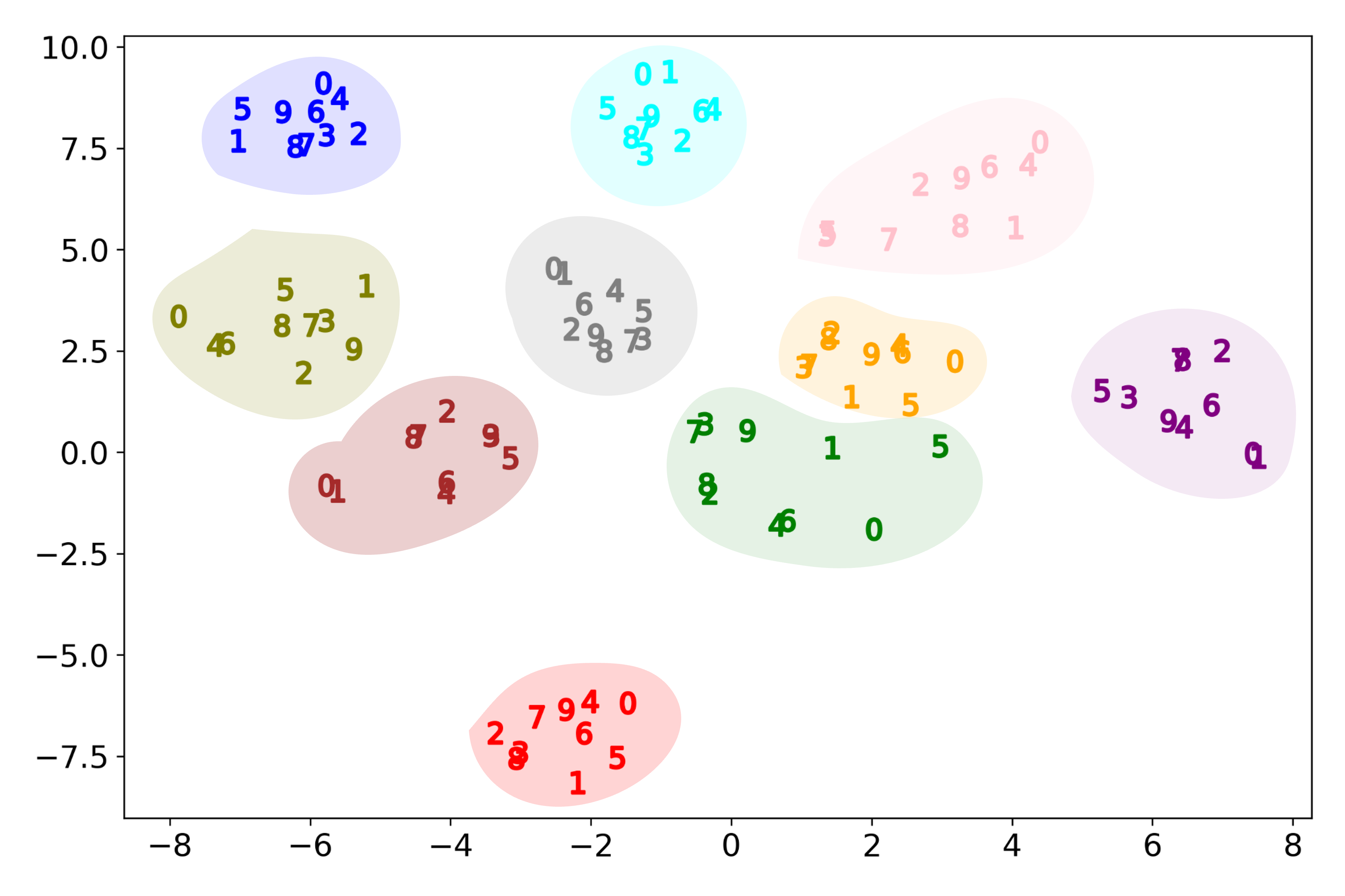}
         \caption{T-SNE visualization of style feature.}
     \end{subfigure}
     \begin{subfigure}{0.48\textwidth}
         \centering
         \includegraphics[width=\textwidth]{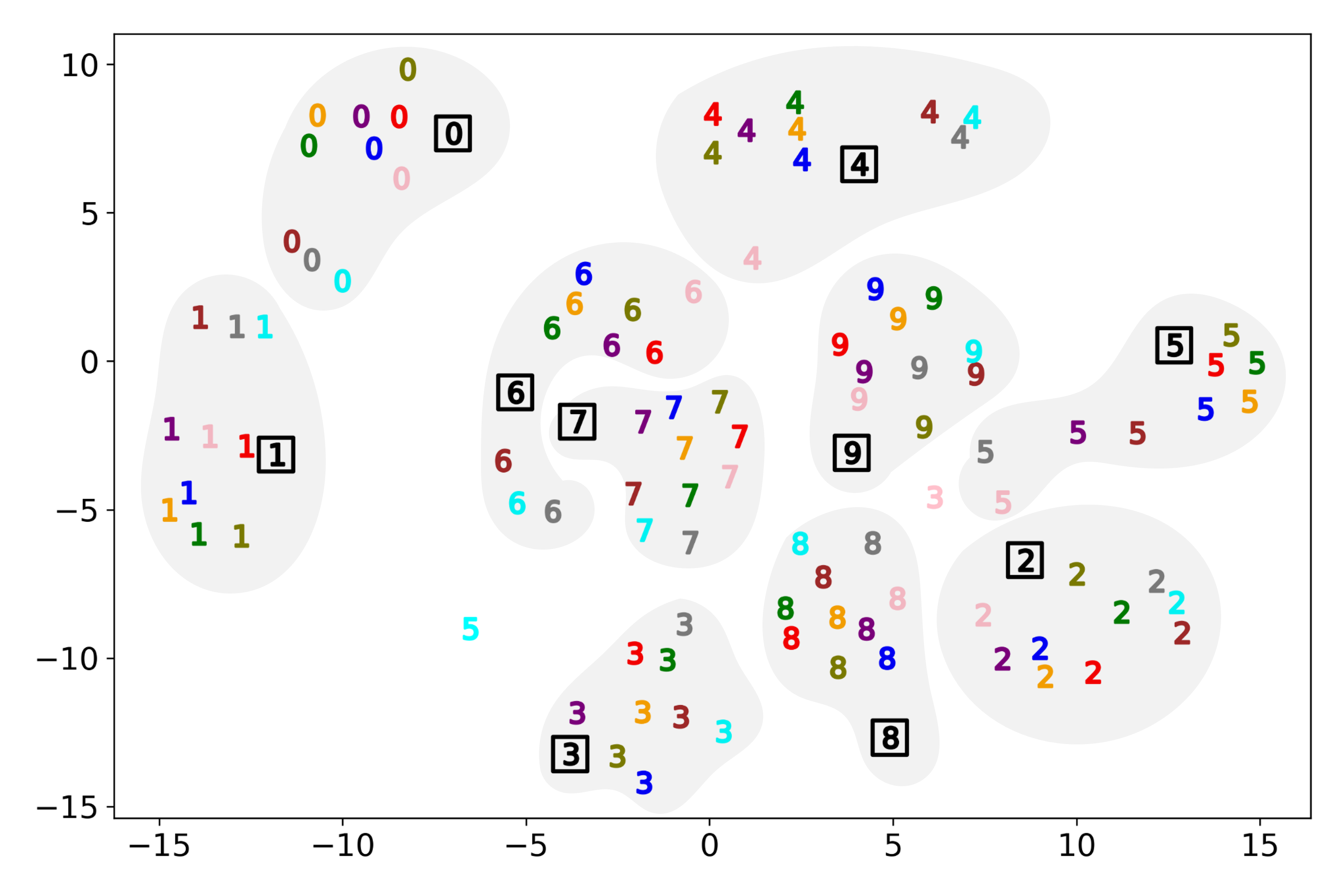}
         \caption{T-SNE visualization of content feature.}
     \end{subfigure}
     
     \caption{(a) shows input images for T-SNE. For the synthesized images, the same column and row indicate the same contents and styles, respectively. Real images of diverse styles are placed on the last row. (b) and (c) present visualizations of decomposed style and content features. The colored numbers indicate synthetic examples including 10 contents (numbers) with 10 styles (colors) and the boxed numbers represent real-world examples.}
     \label{fig:tsne_supp}
\end{figure*}

\subsection{Experiments on Feature Decomposition}

\subsubsection{T-SNE Visualization of Style and Content Features}
We present synthetic and real images, which have been employed for visualization of T-SNE, in Figure~\ref{fig:tsne_supp} (a).
We utilize a rendering tool to achieve synthetic images that have the same styles with different contents and we exploit the IC15~\cite{IC15} as the real images.
Figure~\ref{fig:tsne_supp} (b) and (c) show that the same styles and contents are plotted closely in the style and content feature spaces, respectively. 
Moreover, as shown in Figure~\ref{fig:tsne_supp} (c), we observe similar contents are located closely in the content features space where Content 6 (``SALE'') and Content 7 (``sale'') appear adjacent to each other.



\subsubsection{Critical Components for Generation}
\begin{figure*}[t!]
    \centering
    \includegraphics[width=1\linewidth]{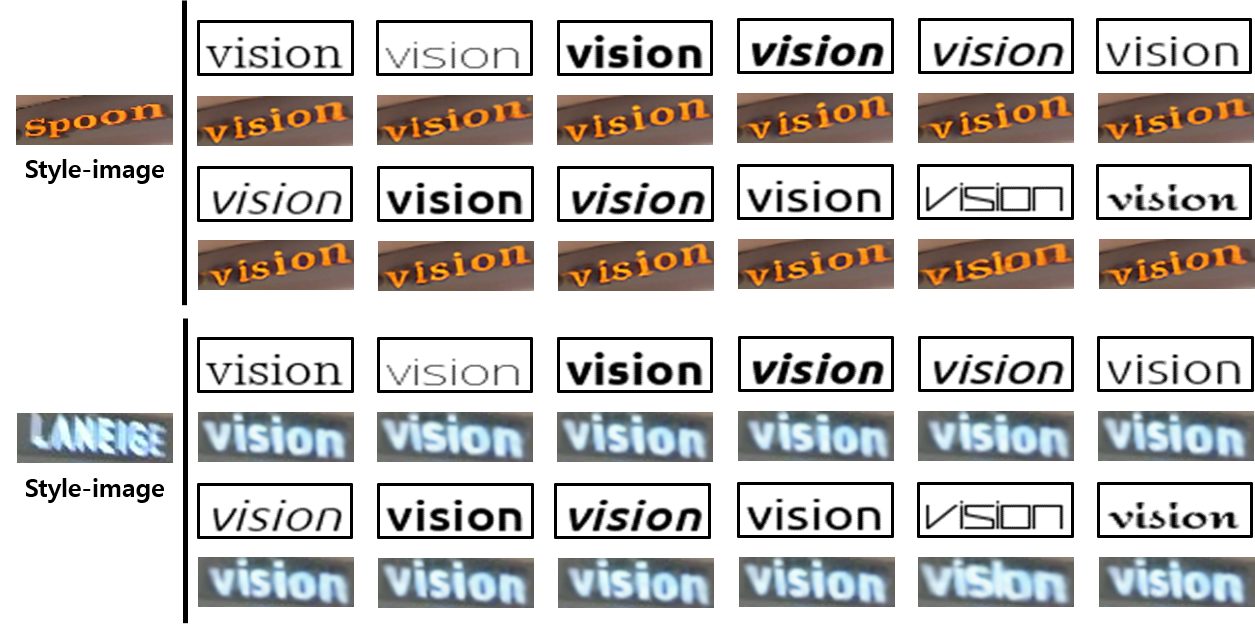}
    \caption{The generated images from RewriteNet with changing the font of content-image. The black rectangular image represents content-image and its corresponding output is listed below.
    }
    \label{fig:feature_decomp_font}

    \centering
    \includegraphics[width=1\linewidth]{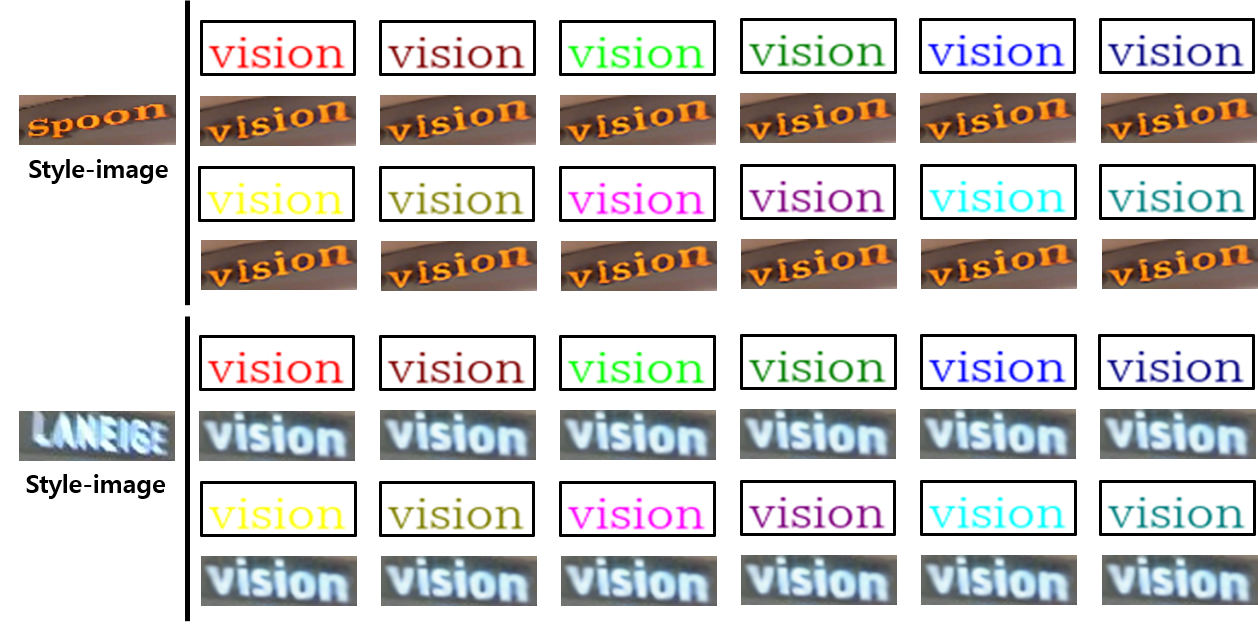}
    \caption{The generated images from RewriteNet with changing the color of content-image. The black rectangular image represents content-image and its corresponding output is listed below.
    }
    \label{fig:feature_decomp_color}
\end{figure*}

We have validated feature decomposition between style and content in Figure 6 of the main manuscript where the generated images are quite stable to the change of content images.
We investigate which component mainly affects the style of the generated image by feeding various fonts and colors of content-image.
As can be seen in Figure~\ref{fig:feature_decomp_font}, the generated images are stable according to the change of font but slightly different from each other.
On the other hand, the generated images are invariant according to the change of color as shown in Figure~\ref{fig:feature_decomp_color}.
We also measure distances between the generated images and variance of the generated images with three metrics: 
\begin{itemize}
    \item PSNR (Peak Signal-to-Noise Ratio): pixel-wise MSE (Mean Squared Error) based distance. Since we cannot achieve the original (reference) image, we measure the distance between generated images. 
    \item SSIM (Structural Similarity): perceptual quality-based distance. Since we cannot achieve the original (reference) image, we measure the distance between generated images. 
    \item Variance: the averaged variance of the generated images.  
\end{itemize}
As can be seen in Table~\ref{table:generated_distance}, the stability of the generated images is more affected by the change of font. 

\begin{figure*}[t!]
    \centering
    \includegraphics[width=1\linewidth]{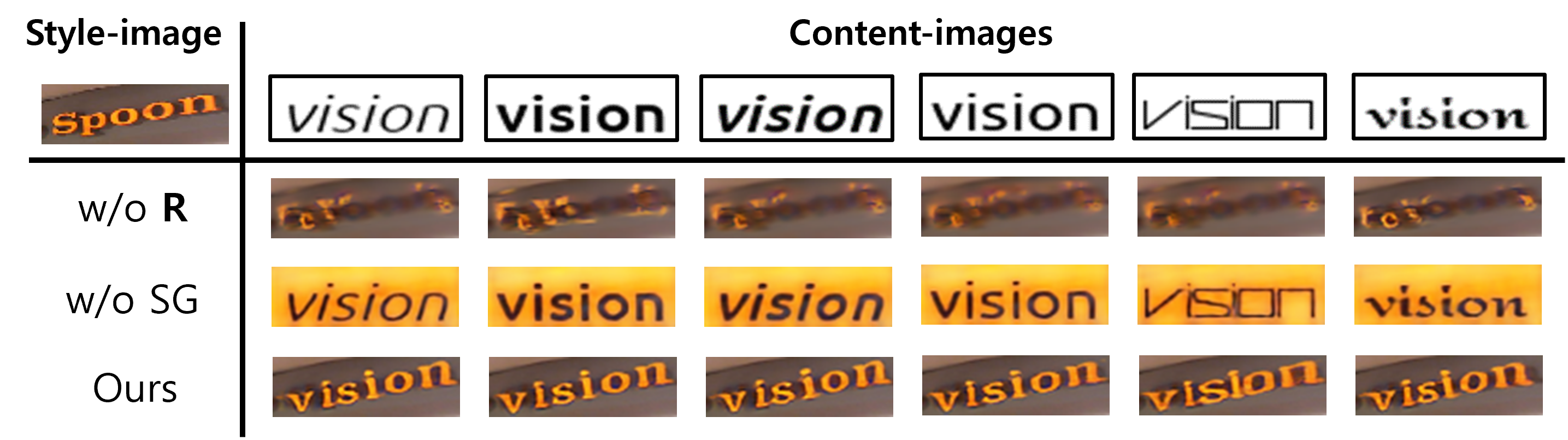}
    \vspace{3.0em}
    \includegraphics[width=1\linewidth]{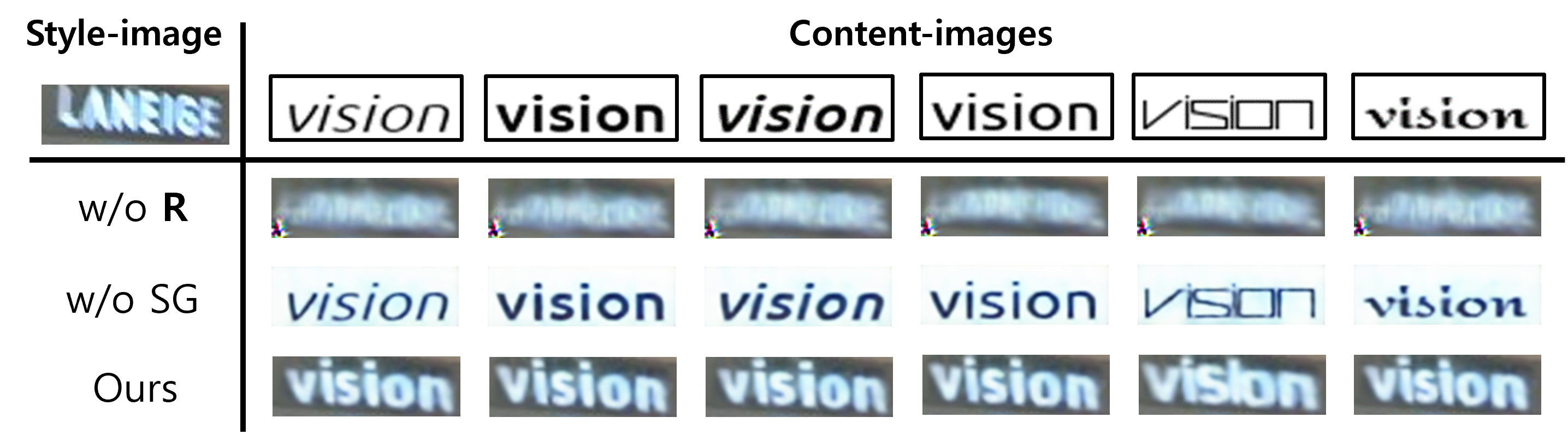}
    \caption{The generated images from ``w/o $\mathbf{R}$'', ``w/o stop gradient'', and RewriteNet with changing the font of content-image. 
    }
    \label{fig:ablation_decomp_font}
\end{figure*}

We also investigate other ablated models, which have been suspected that features are not well separated, by feeding various fonts of content-image.
As shown in Figure~\ref{fig:ablation_decomp_font}, the generated images from ``w/o stop gradient'' significantly vary according to the change of font of content-images. Although ``w/o $\mathbf{R}$'' achieves stable results according to the change of content-images, its results show that content-images are not well employed. 



\begin{table}[t]
\begin{center}
\begin{tabular}{lccc}
\toprule 
\textbf{Changes} & \textbf{PSNR} ($\uparrow$) & \textbf{SSIM} ($\uparrow$) &  \textbf{Var} ($\downarrow$) \\
\midrule \midrule
Font  & 18.51 & 0.6242 & 7.9 $\times 10^{\text{-3}}$\\
Color  & $Inf$ & 0.9942 & 0.08 $\times 10^{\text{-3}}$ \\
\bottomrule
\end{tabular}
\end{center}
\caption{PSNR (dB), SSIM, and the variance between generated images. $Inf$ denotes infinity that indicates some of the generated images are exactly identical.
}
\label{table:generated_distance}
\end{table}

\subsection{Learning from Text Edited Images}
\subsubsection{Training and Evaluation Details}
For the generation of training data, the unified real-world data (59,856 in total), combining four benchmark training datasets such as IIIT~\cite{IIIT5K}, IC13~\cite{IC13}, IC15~\cite{IC15}, and COCO~\cite{COCO}, is used as the style-images.
We generate 18 text images from a single style image. As a result, the total amount of generated images is about 1M (59,856 $\times$ 18). 

We focus on irregular shaped real-world data, because, it is more challenging with diverse curve text alignment that could directly show the style-preservation performance.    
Thus, all trained models are evaluated on the three benchmarks where the total number of images is 2,744; 1,811 from IC15~\cite{IC15}, 645 from SVTP~\cite{SVTP}, and 288 from CT80~\cite{CUTE80} following the evaluation protocol of scene text recognition (STR)~\cite{baek2019STRcomparisons}. 

\subsubsection{Validations on Multiple STR Models}
We will validate the effects of our generated data on different STR models such as CRNN~\cite{CRNN}, RARE~\cite{shi2016robust} and TRBA~\cite{baek2019STRcomparisons}. 
As presented in Table~\ref{table:str_differentmodels_small}, our generated data improves STR performances on all baselines. 
These results show that RewriteNet generates reliable scene text examples, which can be well generalized to multiple STR models.

\begin{table}[t]
\tabcolsep=0.09cm
\centering
\begin{tabular}{cc|l|ccc} 
\toprule
& Model & Train Data & IC15 & CUTE80 & SVTP \\
\midrule
\midrule
&CRNN~\cite{CRNN} & Synth &  	69.8 &	66.3 & 71.3 \\ 
&CRNN~\cite{CRNN} & Synth+Ours & 70.9  & 77.8 & 71.3  \\ 
\midrule
&RARE~\cite{shi2016robust} & Synth  & 75.7  &  73.3 & 75.7  \\ 
&RARE~\cite{shi2016robust} & Synth+Ours &  76.3  & 83.0 & 78.6 \\  
\midrule
&TRBA~\cite{baek2019STRcomparisons} & Synth &   78.0  & 76.7 & 79.5 \\
&TRBA~\cite{baek2019STRcomparisons} & Synth+Ours & 79.6 & 84.4 & 81.6 \\
\bottomrule
\end{tabular}
\caption{Average STR accuracy of three benchmark test datasets depending on the training data.
``Synth'' indicates font-rendered data from MJSynth~\cite{Jaderberg16mjsynth} and SynthText~\cite{Gupta16synthtext}.
``Ours'' represents fully generated data from unlabeled real images using  RewriteNet.}
\label{table:str_differentmodels_small}
\end{table}

\end{appendix}
\end{document}